\documentclass{article}
\usepackage[a4paper,left=2.2cm,right=2.2cm,bottom=1.5cm,top=1.5cm,includefoot,includehead]{geometry}

\usepackage[utf8x]{inputenc} % allow utf-8 input
\usepackage[T1]{fontenc}    % use 8-bit T1 fonts
\usepackage{hyperref}       % hyperlinks
\usepackage{url}            % simple URL typesetting
\usepackage{booktabs}       % professional-quality tables
\usepackage{amsfonts}       % blackboard math symbols
\usepackage{nicefrac}       % compact symbols for 1/2, etc.
\usepackage{microtype}      % microtypography

\usepackage{amsmath}
\usepackage{amssymb}
\usepackage{mynotation}

\usepackage{algorithm}
\usepackage{algorithmic}

\usepackage{xr}

\usepackage{authblk}

\usepackage[pdftex]{graphicx} % more modern
\usepackage{subcaption} 
\usepackage{wrapfig}
\usepackage{import}

\usepackage{tikz}
\usetikzlibrary{arrows,shapes,decorations.pathmorphing}
\usetikzlibrary{backgrounds,calc,positioning,fadings}
\tikzset{>=stealth'} 
\tikzstyle{graphnode} = 
[circle,draw=black,minimum size=22pt,text centered,text
width=25pt,inner sep=0pt] 
\tikzstyle{var}   =[graphnode,fill=white]
\tikzstyle{obs}   =[graphnode,fill=black,text=white]
\tikzstyle{fac}   =[rectangle,draw=black,fill=black!25,minimum size=5pt]
\tikzstyle{facprior} =[rectangle,draw=black,fill=black,text=white,minimum size=5pt]
\tikzstyle{edge}  =[draw=white,double=black,thick,-]
\tikzstyle{prior} =[rectangle, draw=black, fill=black, minimum size=
5pt, inner sep=0pt]
\tikzstyle{dirprior} = [circle, draw=black, fill=black, minimum
size=5pt, inner sep=0pt]

\usepackage{pgfplots}

\usepackage{xcolor}

\usepackage{tcolorbox}

\title{Deep kernel learning for integral measurements}
\author[1]{Carl Jidling}
\author[2]{Johannes Hendriks}
\author[1]{Thomas B. Sch\"{o}n}
\author[2]{Adrian Wills}
\affil[1]{Department of Information Technology, Uppsala University, Sweden}
\affil[2]{School of Engineering, University of Newcastle, Australia}

\date{}

\begin{document}
	
	\maketitle
	
\begin{abstract}
	Deep kernel learning refers to a Gaussian process that incorporates neural networks to improve the modelling of complex functions. 
	We present a method that makes this approach feasible for problems where the data consists of line integral measurements of the target function. 
	The performance is illustrated on computed tomography reconstruction examples.
\end{abstract}

\section{Introduction}

The Gaussian process (GP) \cite{Rasmussen2006} is a powerful regression tool that has been successfully applied to problems within many different fields.
Encoding a broad class of non-linear functions, a key feature of the GP is the ability to adapt its complexity with the size of the data set while keeping a constant number of free hyperparameters; this is referred to as flexibility.  
The performance and accuracy of the GP is, however, in no small part determined by the model assumptions embedded in the associated covariance function.

The most common covariance functions are stationary, which means that the modelled correlation between two function values is dependent purely on the distance between their corresponding input locations.
A notable member of this class is the \textit{squared exponential} covariance function, which is widely employed mainly due to its ease of implementation.

Although stationary covariance functions are intuitive and rather realistic for many functions, this choice causes severe problems if the target function contains non-smooth features, such as rapid, step-like changes.
A way of meeting this challenge is to use a non-stationary covariance function. 
In particular, the \textit{neural network} covariance function \cite{Neal1996} is known for its ability to capture non-stationary features.
However, it is harder to implement and extend beyond point measurement models (direct observations of the target function), e.g. extension to integral measurements.
\begin{figure}[hb]		
	\begin{center}
		%	\vskip -0.3in
		%	\vspace{0mm}
%		\begin{tcolorbox}
%			\red{cheese figure}
%			\vspace{4cm}
%		\end{tcolorbox}
		\centerline{\includegraphics[width=\textwidth]{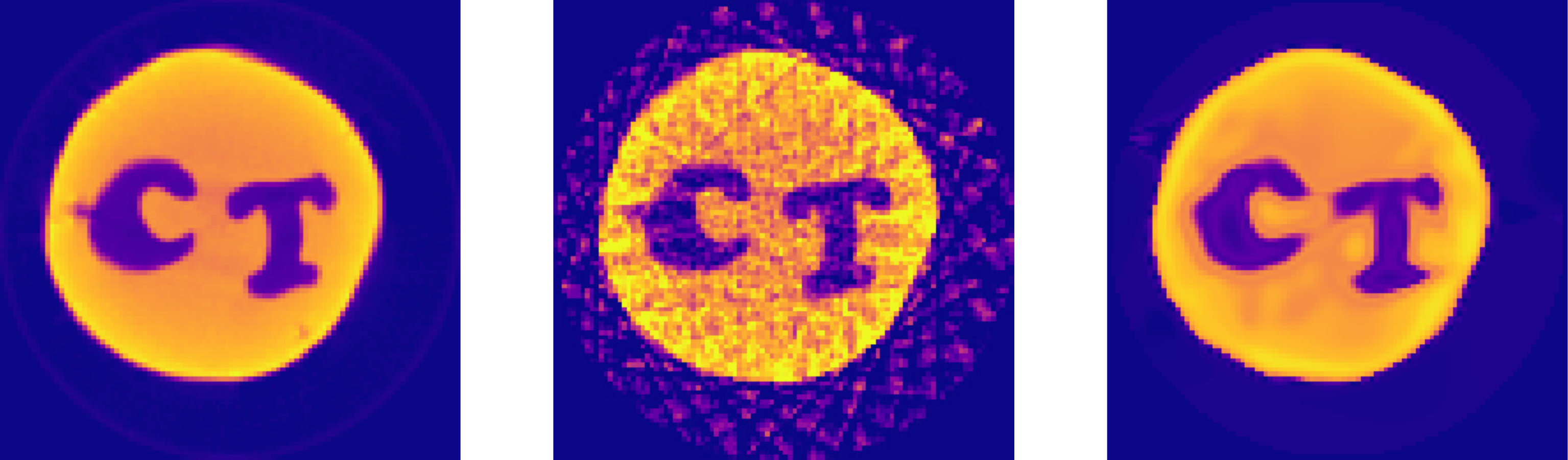}}
			\vspace{-2mm}
		\caption{The proposed method applied to computed tomography reconstruction from X-ray data. The following example is from the scanning of a carved cheese. Left: reference reconstruction of high accuracy. Middle: reconstruction using filtered back projection, a commonly used method in practise. Right: reconstruction using the proposed method.}
		\label{fig:cheese}
	\end{center}
	\vskip -0.3in
\end{figure}

An alternative non-stationary construction is obtained by warping the inputs to a stationary covariance function through a non-linear mapping \cite{Rasmussen2006}.
Letting this mapping be a neural network, we obtain the framework known as \textit{manifold Gaussian processes} \cite{Calandra2016} or \textit{deep kernel learning} \cite{Wilson16}, which has been demonstrated on point measurements with promising potential.

The practical procedure of this approach becomes more challenging when we consider more advanced measurement models.
In this work, we propose a method that allows for the application of deep kernel learning to problems where the measured data is expressed as line integrals of the target function, which arise for instance within X-ray computed tomography (CT) \cite{Shepp1978computerized,Herman1979image,Cormack1963representation} and strain field estimation \cite{Hendriks2017,Wensrich2016b,Lionheart2015}.
This is a non-trivial extension considering that a straightforward naive implementation requires numerical double integral computations in a number that scales quadratically with the size of the data set.
Also, the training procedure is challenging since the cost function contains many local minima and the convergence is dependent upon a suitable initialisation. 

To overcome these obstacles we approximate the GP with a Hilbert space basis function expansion \cite{SolinSarkka}, and so reduce the numerical computation to \textit{single integrals} in a number that scales \textit{linearly} with the size of the data set.
Furthermore, by exploiting the model setup we pre-train the neural network used in the covariance function to provide a customised initialisation for the remaining joint training when incorporated within the deep kernel model. This significantly improves the end result and overall robustness.

The potential is demonstrated on simulated and real-data CT reconstruction problems, with promising results shown.
An illustration is seen in Figure \ref{fig:cheese}.

\section{Background on the model}
In this section, we briefly introduce the model background that forms the foundation of our developments; integral measurements in GPs and the deep kernel learning formulation. 

\subsection{Gaussian processes with integral measurement}
The GP can be seen as a distribution over functions where any finite set of function values has a joint Gaussian distribution.
Formally we write
\begin{align}
f(\x)\sim\GPd{0}{k(\x,\x')},
\end{align}
to denote that the function $f(\x):\Rn^{D_\x}\rightarrow\Rn$ is modelled as a zero-mean GP with covariance function $k(\x,\x'):\Rn^{D_\x}\times\Rn^{D_\x}\rightarrow\Rn$, and $\x=[x_1,\dots,x_{D_\x}]^\Transp\in\Rn^{D_\x}$.

An important and very useful property of the GP is that it is closed under linear functional evaluations \cite{Papoulis1991,Rasmussen2006,Garnett2017,Jidling2017}.
This means that when a linear functional $\Lx$ is acting on a GP, the result is also a GP.
Hence, it holds that
\begin{align}
\Lx f(\x) \sim \GPd{0}{\Lx \Lxp k(\x,\x')},
\end{align} 
where $\Lxp$ denotes the functional acting on the second argument of $k(\x,\x')$.
Considering line integrals along straight line segments, we define the functional as
\begin{align}
\Lx_i f(\x) \triangleq \int_{-r_i}^{r_i} f(\x^0_i + s\nh_i)ds,
\end{align}
where $\x^0_i$ denotes the centre of the line, $\nh_i$ is a unit vector specifying the direction, $r_i$ is the integration radius (half the line length) and the index $i$ refer to the $i^\text{th}$ data point.
The corresponding covariance transformation yields the double integral
\begin{align}\label{eq:double_int}
\Lx_i \Lxp_j k(\x,\x') = \int_{-r_i}^{r_i}\int_{-r_j}^{r_j} k(\x^0_i+s\nh_i,\x^0_j+s'\nh_j)dsds',
\end{align}
which gives the covariance between measurement $i$ and $j$.
Let the measurements be stored in the vector $\y=[y_1,\dots,y_N]^\Transp$ with
\begin{align}\label{eq:meas_eq}
y_i = \Lx_i f(\x)+\varepsilon_i,
\end{align}   
where the noise $\varepsilon_i\sim\Nd{0}{\sigma^2}$.
Furthermore, we are interested in the prediction $\f_*=~[f(\x_{*1})\cdots f(\x_{*N_*})]$, the function values at a set of unseen input locations $\{\x_{*i}\}_{i=1}^{N_*}$.
Since linear transformations preserve Gaussianity, $\y$ and $\f_*$ have a joint Gaussian distribution:
\begin{equation} \label{eq:joint_func}
\begin{bmatrix}
\y \\
\f_*       
\end{bmatrix}
\sim\mathcal{N}
\Bigg(
\begin{bmatrix}
\vec{0} \\
\vec{0}
\end{bmatrix},
\begin{bmatrix}
\Lcal+\sigma^2I & \Lcal_*\\
\Lcal_*^\Transp& K_{**}
\end{bmatrix}
\Bigg),
\end{equation} 
where $\Lcal_{ij}=\Lx_i\Lxp_j k(\x,\x')$, $(\Lcal_*)_{ij}=\Lx_i k(\x,\x_{*j})$, and $(K_{**})_{ij}=k(\x_{*i},\x_{*j})$.
The predictive expressions given this joint prior are given by
\begin{subequations} \label{eq:gp_pred}
\begin{align} 
\E[\f_*|\y] &= \Lcal_*^\Transp \Lcal^{-1} \y , \label{eq:gp_pred_mean}\\
\Cov{ \f_*|\y }  &= K_{**} -  \Lcal_*^\Transp \Lcal^{-1} \Lcal_*. \label{eq:gp_pred_cov}
\end{align}
\end{subequations}
Thus, we can make predictions of the function values $\f_*$ purely based on line integral data; note that integration is a \textit{conservative} functional, meaning that all information about the function is preserved under its evaluation (as opposed to e.g. differentiation).
The challenging part here lies in the computation of the integral expressions, especially the double integrals \eqref{eq:double_int} in $\Lcal$.

\subsection{Deep kernel learning}
The most crucial part of Gaussian process modelling is the selection of the covariance function $k(\x,\x')$, since it stipulates the basic behaviour of the target function $f(\x)$.
The most common covariance functions are stationary such that $k(\x,\x')=k(\x-\x')$.
Prominent members of this class include the Mat\'{e}rn family \cite{Stein1999}, the so-called spectral mixture kernels \cite{Wilson13}, and the popular squared exponential covariance function:
\begin{align}
\label{eq:se}
k(\x,\x')=\sigma_f^2\exp\left[-\frac{1}{2}\sum_{k=1}^{D_{\x}}l_k^{-2}(x_k-x_k')^2\right],  
\end{align}
parameterised by the magnitude parameter $\sigma_f$ and the lengthscales $l_k$, which impact how quickly the function may change.

In order to extend the expressiveness of stationary covariance functions, non-stationarity can be introduced by transforming the inputs through a non-linear mapping $\u(\cdot):\Rn^{D_\x}\rightarrow \Rn^{D_\u}$ to form $k(\u(\x),\u(\x'))$ \cite{Rasmussen2006}.
%The non-linearity implies that $k_\text{new}(\cdot,\cdot)$ becomes non-stationary although $k(\cdot,\cdot)$ is stationary. 
The dimension $D_\u$ of $\u(\cdot)$ can be chosen arbitrarily, and may therefore differ from the dimension $D_\x$ of $\x$.
%, so the warped equivalent to \eqref{eq:se} is
%\begin{align}
%\label{eq:se_warped}
%k(\x,\x')=\sigma_f^2\exp\left[-0.5\sum_{k=1}^{D_\u}l_j^{-2}(u_k(\x)-u_k(\x'))^2\right]. 
%\end{align}

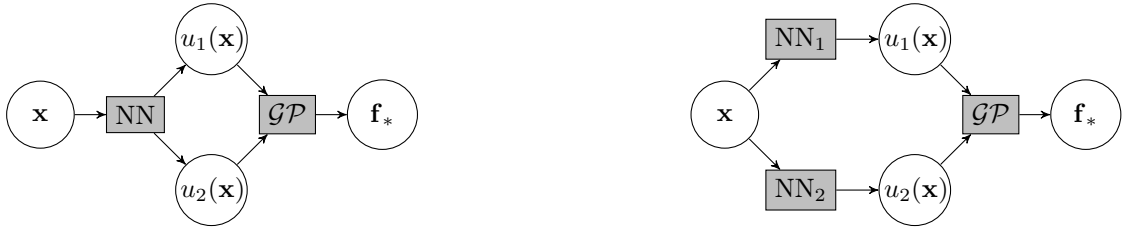
\begin{figure}
	\begin{minipage}[b]{.45\textwidth}
	\centering
	\begin{tikzpicture}
	\node[var] at (-0.25,0) (x) {$\x$};
	\node[fac] at (1,0) (NN) {NN};
	\draw [->] (x) -- (NN);
	\node[var] at (2,1) (u1) {$u_1(\x)$};
	\draw [->] (NN) -- (u1);
	\node[var] at (2,-1) (u2) {$u_2(\x)$};
	\draw [->] (NN) -- (u2);
	\node[fac] at (3,0) (GP) {$\GP$};
	\draw [->] (u1) -- (GP);
	\draw [->] (u2) -- (GP);
	\node[var] at (4.25,0) (fstar) {$\f_*$};
	\draw [->] (GP) -- (fstar);
	\end{tikzpicture}
	\end{minipage}
	\hfill
	\begin{minipage}[b]{.45\textwidth}
	\centering
	\begin{tikzpicture}
	\node[var] at (0,0) (x) {$\x$};
	\node[fac] at (1,1) (NN1) {NN$_1$};
	\draw [->] (x) -- (NN1);
	\node[fac] at (1,-1) (NN2) {NN$_2$};
	\draw [->] (x) -- (NN2);
	\node[var] at (2.5,1) (u1) {$u_1(\x)$};
	\draw [->] (NN1) -- (u1);
	\node[var] at (2.5,-1) (u2) {$u_2(\x)$};
	\draw [->] (NN2) -- (u2);
	\node[fac] at (3.5,0) (GP) {$\GP$};
	\draw [->] (u1) -- (GP);
	\draw [->] (u2) -- (GP);
	\node[var] at (4.75,0) (fstar) {$\f_*$};
	\draw [->] (GP) -- (fstar);
	\end{tikzpicture}
	\end{minipage}
	\caption{Illustration of deep kernel learning and two different constructions for the latent mapping $\u(\x)=[u_1(\x)\,\,\,u_2(\x)]^\Transp$. Left: the components $u_1(\x)$ and $u_2(\x)$ are different outputs of the same neural network. Right: the components are independent outputs of two different networks.}	
%	\vspace{-2mm}
	\label{fig:latent_ill}
\end{figure}

Using this construction in the modelling of complex functions with limited prior knowledge, we need $\u(\cdot)$ to encode a general class of functions that can be learnt from data.
A natural choice is to let $\u(\cdot)$ be described by a neural network.
This is the idea behind manifold GPs \cite{Calandra2016} and deep kernel learning \cite{Wilson16}. %, with the main difference being that the latter procedure includes an approximative covariance model for improved scalability.
The $D_\u$ latent outputs $u_j(\cdot)$ are either completely independent, or they are different outputs of the same network, see Figure \ref{fig:latent_ill} for an illustration.
The intuition is that the neural network does not have to learn the complete function $f(\x)$, but only identify its discontinuities while for the remaining part the model can rely upon the regression capabilities of the GP.

\section{Deep kernel learning with integral measurements}
Our aim in this work is to combine the GPs ability of incorporating line integral measurement with the neural network warping to form a method that is practically feasible beyond one-dimensional problems. 

\subsection{Basis function expansion}
To reduce the computational load, we make use of a Hilbert space approximation method for GP regression \cite{SolinSarkka}. 
In this approach a stationary covariance function is approximated by the following finite sum:
\begin{equation}\label{eq:basis_sum}
k(\u,\u')\approx \sum_{j=1}^{m}S(\cc_j) \phi_j(\u)\phi_j(\u'),
\end{equation}
where $S(\cdot)$ denotes the spectral density of the covariance function. 
The basis functions $\{\phi_j(\cdot)\}_{j=1}^m$ with corresponding eigenvalues $\{\lambda_j\}_{j=1}^m$ are obtained from the Laplace eigenvalue problem 
\begin{align}\label{eq:Laplace}
\begin{cases}
\hspace{-4mm}
\begin{split}
-\Delta\phi_j(\u)&  = \lambda_j\phi_j(\u),  \\[2mm]
\phi_j(\u)&         =0,         
\end{split}
\end{cases}
\quad
\begin{split}
\u&\in\Omega, \\
\u&\in\partial\Omega, 
\end{split}
\end{align}
where $\Omega=[L_1,L_1]\times\cdots\times[L_{D_\u},L_{D_\u}]$ is a generalised rectangular domain, and $\Delta$ denotes the Laplace operator. % $\Delta\phi=\nabla\cdot\nabla\phi$. 
Here a Dirichlet boundary condition is used, but it does not affect the GP solution if $L_k$ is chosen carefully, which is discussed in the supplementary material; for problems where the boundary conditions are explicitly specified, more advanced formulations are possible \cite{Solin_kok_bc}. 
The solution to \eqref{eq:Laplace} is given by
\begin{align}\label{eq:basis_funcs}
\phi_j(\u) = \prod_{k=1}^{D_\u}L_k^{-1/2} \sin\left[ c_{kj}(u_k+L_k)\right], 
\qquad
c_{kj} = \frac{j_k\pi}{2L_k}, 
\qquad
\lambda_j = \sum_{k=1}^{D_\u} c_{kj}^2.
\end{align}
The notation denotes that basis function $j$ has index $j_k\in[1,\dots,\tilde{m}]$ in direction $k$.
Using $\tilde{m}$ basis functions in each direction, we get a total number of $m=\tilde{m}^{D_\u}$.
Furthermore, we have introduced $\cc_j=[c_{1j},\dots,c_{D_\u j}]$ for the input to $S(\cdot)$ in \eqref{eq:basis_sum}.

With the network warping included, the matrix $\Lcal$ in \eqref{eq:joint_func} is approximated as $\Lcal\approx\Phi\Lambda\Phi^\Transp$ where
\begin{align}
\Phi_{ij} = \Lx_i \phi_j(\u(\x)), \qquad
\Lambda_{jj} = S(\cc_j).
\end{align} 
Using the matrix inversion lemma, the approximate versions of the predictive expressions \eqref{eq:gp_pred} are reformulated for more efficient computations provided that $m<N$. 
However, the main advantage of this method is the separation of the inputs $\x$ and $\x'$ in the basis function product.
A consequence of this separation is that the double integral computations required to build $\Lcal$ \emph{reduce to single} integral computations of the form
\begin{align}
\Phi_{ij} = \int_{-r_i}^{r_i} \prod_{k=1}^{D_\u}L_k^{-1/2} \sin\left[ c_{kj}(u_k(\x^0_i+s\nh_i)+L_k) \right] ds.
\end{align}
This integral can not be computed in closed form due to the non-linearity $\u(\cdot)$, but numerical integration is nevertheless significantly less demanding in one dimension than it is in two (compare with expression \eqref{eq:double_int}). 
For instance, we can use a direct scheme % of the form
%\begin{align}
%\Phi_{ij} \approx \sum_{q=0}^{n_i} w_q \prod_{k=1}^{D_\u}L_k^{-0.5} \sin\left[ c_{kj}(u_k(\x^0_i+(-r_i+q\Delta_s)\nh_i)+L_k) \right],
%\end{align}
such as the composite Simpson's $1/3$ rule of integration \cite{Chapra}. 

\subsection{Training the model}\label{sec:training}
The model as formulated above contains the free hyperparameters $\theta=[\theta_\text{k}^\Transp\,\,\,\theta_\u^\Transp]^\Transp$, which we separate with respect to the covariance function ($\theta_\text{k}$) and the neural network ($\theta_\u$), respectively.
As for the squared exponential covariance function \eqref{eq:se}, we have $\theta_\text{k}=\{\sigma_f,\{l_k\},\sigma\}$, including the standard deviation $\sigma$ of the noise.
There are different cost function options available for training $\theta$, among which two common ones are the marginal likelihood (ML) and leave-one-out cross-validation (LOO-CV) \cite{Rasmussen2006}.
In our experience, these two methods have shown a similar performance.  
Regardless of choice, the numerical robustness of the computations is improved using the $QR$-factorisation with details given in the supplementary material.

An important aspect of the training procedure is the parameter initialisation in the optimisation routine.
This is a non-trivial challenge as the total number of parameters is large due to the neural network.
Moreover, the initialisation typically has big impact on the resulting optimisation performance and hence also on the quality of the final prediction. 
For challenging problems with complex two-dimensional functions, pre-training of the neural network as described below has shown to have a crucial impact on the convergence.

To obtain a satisfying initial guess, let us take a moment to reflect on what we want $\u(\cdot)$ to achieve. 
The reason for introducing this mapping is that a stationary covariance function always assigns high correlation to function values at closely located inputs.
In other words, if the distance $|\x-\x'|$ is small, then $f(\x)$ and $f(\x')$ are assumed to be similar. 
In regions of rapid changes and discontinuities, this assumption fails drastically. 
We concretise this by considering inference of a one-dimensional step function while using a scalar latent mapping $u(\cdot)$.
Two points $x_0$ and $x_1$ located just before and just after the step differ significantly in their function values $f(x_0)$ and $f(x_1)$.
Therefore, we want to train $u(\cdot)$ such that $u(x_0)$ and $u(x_1)$ become clearly separated, and hence make $f(x_0)$ and $f(x_1)$ weakly correlated.
Considering the remaining parts of the step function, it consists of two constant regions where we also want $u(\cdot)$ to be constant for maximum correlation.

Extending this reasoning, it is easy to imagine several different mappings that would yield ideal correlation assignments by the stationary covariance function, with the essential feature being identification of discontinuities and distinguishing between points that are separated by them. 
An intuitive ideal mapping is $\u(\x)$ being equal to the target function, since this choice assigns maximum correlation to identical function values.
Although other ideal mappings might be less complex and more robust, this one is natural in lack of other prior information.
Thus, our proposed pre-training aims at finding a latent mapping that is a reasonable approximation of the true function.

To begin with, we restrict ourselves to the case $D_\u=1$ for now where we denote $\u(\x)$ with $u(\x)$.
Although the model might be more expressive with several latent outputs, the computational load increases since it requires more basis functions; the number scales exponentially with $D_\u$ for a retained frequency resolution.

%Two options for pre-training the neural network parameters $\theta_u$ are presented here.
%The first one is through minimisation of the mean squared error loss directly against the measured data:
%\begin{align}\label{eq:pretrain_direct}
% \theta_u = \underset{\theta_u}{\text{argmin}}	\frac{1}{N}\sum_{i=1}^N (y_i-\Lx_i u(\x))^2,
%\end{align}
%where the integrals $\Lx_i u(\x)$ are computed numerically.
%A second option is given by
%\begin{align}\label{eq:pretrain_gp}
%	\theta_u = \underset{\theta_u}{\text{argmin}} \frac{1}{N_t}\sum_{i=1}^{N_t} (f_t(\x_i)-u(\x_i))^2,
%\end{align}
%where $\{\x_t\}_{i=1}^{N_t}$ is a set of $N_t$ points in the domain of interest, and $f_t(\cdot)$ denotes the solution obtained from standard GP prediction.
%
%Although it is not obvious which option is better, they both give rise to natural questions.
%When \eqref{eq:pretrain_direct} is used, one might wonder what the purpose is of the outer GP layer; it may seem like the training in \eqref{eq:pretrain_direct} can produce a satisfying reconstruction of $f(\x)$ directly.
%Our answer is two-folded: firstly, since the GP prediction is capable of improving upon the latent mapping, combining the ideas yields more robustness than relying upon the network solely; secondly, the GP model is naturally incorporating and estimating the noise level, and does as well provide an uncertainty estimation that a frequentistic technique is lacking.

For pre-training the neural network parameters $\theta_u$, we suggest the choice
\begin{align}\label{eq:pretrain_gp}
\theta_u = \underset{\theta_u}{\text{argmin}} \frac{1}{N_t}\sum_{i=1}^{N_t} (f_t(\x_i)-u(\x_i))^2,
\end{align}
where $\{\x_t\}_{i=1}^{N_t}$ is a set of $N_t$ points in the domain of interest, and $f_t(\cdot)$ denotes the mean prediction obtained from standard GP reconstruction.
As stated, this approach does not directly generalise to the case $D_\u>1$.
However, it can still be employed in such constructions. 
For instance, one could combine a pre-trained neural network mapping $u(\x)$ with the mappings $u_k(\x)=x_k$ (whereby the original inputs are also used). 

%Although it is not obvious which option is better, they both give rise to natural questions.
%When \eqref{eq:pretrain_direct} is used, one might wonder what the purpose is of the outer GP layer; it may seem like the training in \eqref{eq:pretrain_direct} can produce a satisfying reconstruction of $f(\x)$ directly.
%Our answer is two-folded: firstly, since the GP prediction is capable of improving upon the latent mapping, combining the ideas yields more robustness than relying upon the network solely; secondly, the GP model is naturally incorporating and estimating the noise level, and does as well provide an uncertainty estimation that a frequentistic technique is lacking.

A natural question following this pre-training is why we need the neural network; we could as well remove this intermediate step and fix $u(\cdot)$ to be the output of the standard GP. 
However, a standard GP prediction is likely to contain undesired artefacts for problems with discontinuous features, and these artefacts may have negative impact when propagated through to another GP.
With a neural network warping, the joint training is capable of eliminating or at least drastically reduce any impact of that form.   

As for the implementation we make use of PyTorch \cite{paszke2017automatic}, which provides a powerful platform for neural network models.
Employing a gradient-based optimisation routine, we need to compute the partial derivatives of the cost function.
This requires an application of the chain rule, which may not be trivial due to the matrix operations and numerical integration involved; to this end we rely upon PyTorch's support for automatic differentiation.
 
Additionally, a complementary routine for back propagation of derivatives through the $QR$-factorisation has been implemented based on \cite{WalterPHD}, with details described in the supplementary material.
Furthermore, we are using the L-BFGS-optimiser \cite{ShiLBFGS}, modified to allow for a dynamically changing learning rate.

The procedure is summarised in Algorithm \ref{alg:dkl_im}.

\begin{algorithm}[tb]
	\caption{Deep kernel learning with line integral measurements}
	\label{alg:dkl_im}
	\begin{algorithmic}
		\STATE {\bfseries Input:} Data set $\{y,\x^0,r,\nh\}_{i=1}^N$%, prediction point $\{\x_{*i}\}_{i=1}^{N_*}$
		\STATE {\bfseries Output:} $\E[\f_*|\y], \Cov{\f_*|\y}$
		\STATE  {\bfseries 1.} Pre-train the neural network $u(\x)$ using \eqref{eq:pretrain_gp}.
		\STATE  {\bfseries 2.} Train the extended model.
		\STATE {\bfseries 3.} Compute the mean prediction $\E[\f_*|\y]$ and the covariance $\Cov{\f_*|\y}$ using \eqref{eq:gp_pred}. 
		%	\STATE {\bfseries Step 5:}
	\end{algorithmic}
\end{algorithm}

\section{Experimental results}
Here we illustrate the practical performance of the method, starting with a one-dimensional toy example and proceeding with more realistic CT examples. 

\subsection{One-dimensional toy example}

\begin{figure}[!h]
%	\centering
	\begin{minipage}[b]{.45\textwidth}
		\centering
%		\begin{tcolorbox}
%			\red{1D figure}
%			\vspace{4cm}
%		\end{tcolorbox}
		%  
	\scalebox{0.4}{\input{./figures/step.pgf}}

%		\includegraphics{./figures/steppiece.pgf}
%		\caption{One dimensional toy example of inferring a piece-wise constant function from 150 integral measurements over randomly chosen intervals in $[0,1]$. True function in black, standard GP in red and the combined model in blue. The shades indicate the 95$\%$ credibility regions, which is much tighter for the combined model.}
		\caption{One-dimensional toy example of inferring a step function from $50$ integral measurements over randomly chosen intervals in $[0,1]$. True function in solid grey, standard GP in blue (dashed) and the proposed method in red (dash-dotted). The shades indicate the $95\%$ credibility regions, which is much tighter for the proposed method.}
		\label{fig:toy_1D}
	\end{minipage}
	\hfill
	\begin{minipage}[b]{.45\textwidth}
		\centering
		\includegraphics[width=0.65\textwidth]{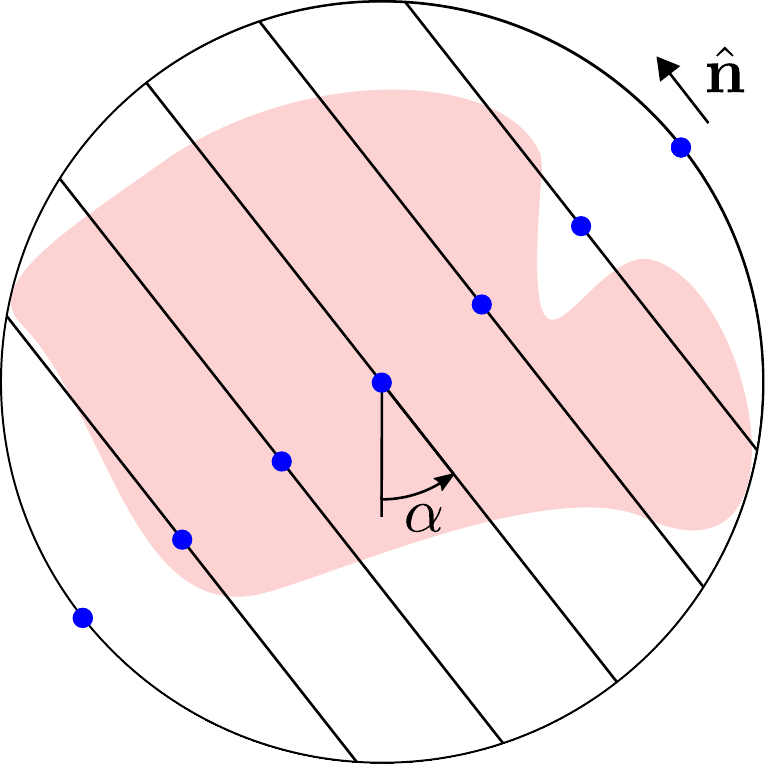}
		\caption{Measurement setup in X-ray computed tomography; illustration of a projection taken from the angle $\alpha$, with the target object shaded. All lines in the projection share the same unit vector $\nh$, while the centre points are different (blue dots). The integration radius $r_i$ is the distance on the line from the centre point $\x_i^0$ to the circle.}
		\label{fig:projection}
	\end{minipage}
\end{figure}

To illustrate the method, we consider inference of the one-dimensional step function seen in Figure~\ref{fig:toy_1D}.
%\begin{align}
%	f(x) = 
%	\begin{cases}
%	1,\qquad x\in[0.2,0.4]\cup[0.6,0.8], \\
%	0, \qquad \text{otherwise.}
%	\end{cases}
%\end{align}
The data set consist of $50$ integrals computed over randomly chosen intervals in the domain $[0,1]$, contaminated by Gaussian noise with standard deviation $0.001$. 
For the latent mapping we are using a neural network with four layers and $(1,5,4,1)$ neurons, employing the hyperbolic tangent activation function after the two hidden layers.
Furthermore, we are using $N_t=100$ uniformly spaced points in the pre-training  \eqref{eq:pretrain_gp}.
The squared exponential covariance function \eqref{eq:se} is used in both the standard GP and the proposed method. 

Figure \ref{fig:toy_1D} shows the result of the proposed method (red dashed-dotted) and the standard GP (blue dashed), together with their $95\%$ credibility regions.
Obviously, the standard GP suffers from its embedded smoothness assumptions and it is struggling with the step, which is reflected in the oscillations and the wide credibility region. % stems from an overestimated noise of $0.020$. 
It should be stressed that this problem differs notably as opposed to considering point measurements from a smooth function -- with complexity added in both the measurement model and the function itself, it becomes significantly more challenging.
Nevertheless, the proposed method performs clearly better than the standard GP, obtaining a good estimate of the true function. 

%with a more accurate noise estimate of $0.0012$. 

\subsection{Computed tomography experiments}
Here, we test the performance on two-dimensional CT problems.
CT provides a good demonstration for our method as it involves line integral measurements of a quantity that can have discrete or sharp changes.  
We compare our proposed method against the filtered back projection (FBP) algorithm.
For decades FBP has served as a state-of-the-art method, in no small part due to the fact that it outperforms iterative optimisation-based alternatives in terms of computation time.
However, FBP is sensitive to noise and demonstrates a relatively poor performance for small data sets -- also referred to as \textit{limited data}. % -- as the problem then becomes ill-posed.
The limited data problem is interesting for several reasons, including: keeping the radiation doses small; efficient use of scanning devices; geometric setup restrictions (as in mammography).  

%Plenty of work has been done on the limited data problem throughout the years, including methods from statistics and deep learning.
In the practical scanning procedure, the data is collected as a set of \textit{projections}, each of which defines a number of parallel lines sharing the same projection angle $\alpha$.
The projection width is determined by the maximum object width $w_\text{max}$, so it is deduced that the entire object is located within a circle of radius $w_\text{max}/2$.
Exploiting this knowledge, the integration radii are found by identifying the intersections between the circle and the straight lines defined by the centre points and the unit vector; see Figure \ref{fig:projection} for an illustration of the geometry. 

In both the examples presented in this section, the neural networks have five layers with $(2,30,20,6,1)$ neurons and the hyperbolic tangent as activation function after the three hidden layers.
Note that this structure is far from optimal and could most likely be improved with a more careful design.
The input domain is normalised to $[-1,1]\times[-1,1]$, and the pre-training is using $N_t=10^4$ uniformly spaced points.
Also, all GPs are using the squared exponential covariance function \eqref{eq:se}.
We compare the results to FBP reconstructions computed with the \texttt{iradon} command from the \texttt{skimage} module in Python \cite{scikit-image}.
The simulated data is generated with the corresponding \texttt{radon} command using a high-resolved version of the ground truth image.

\subsubsection{Simulated data}
As a simulated example we consider the Shepp-Logan phantom \cite{SheppLogan}.
The data consist of $9$ projections evenly spaced in $[0,160]^\circ$ with $185$ lines each, yielding a total of $1\,665$ measurements.
Furthermore, Gaussian noise with standard deviation $0.001$ is added on top.

\begin{figure}[hb]		
	\begin{center}
		%	\vskip -0.3in
		%	\vspace{0mm}
%		\begin{tcolorbox}
%			\red{phantom figure}
%			\vspace{4cm}
%		\end{tcolorbox}
		\centerline{\includegraphics[width=1\textwidth]{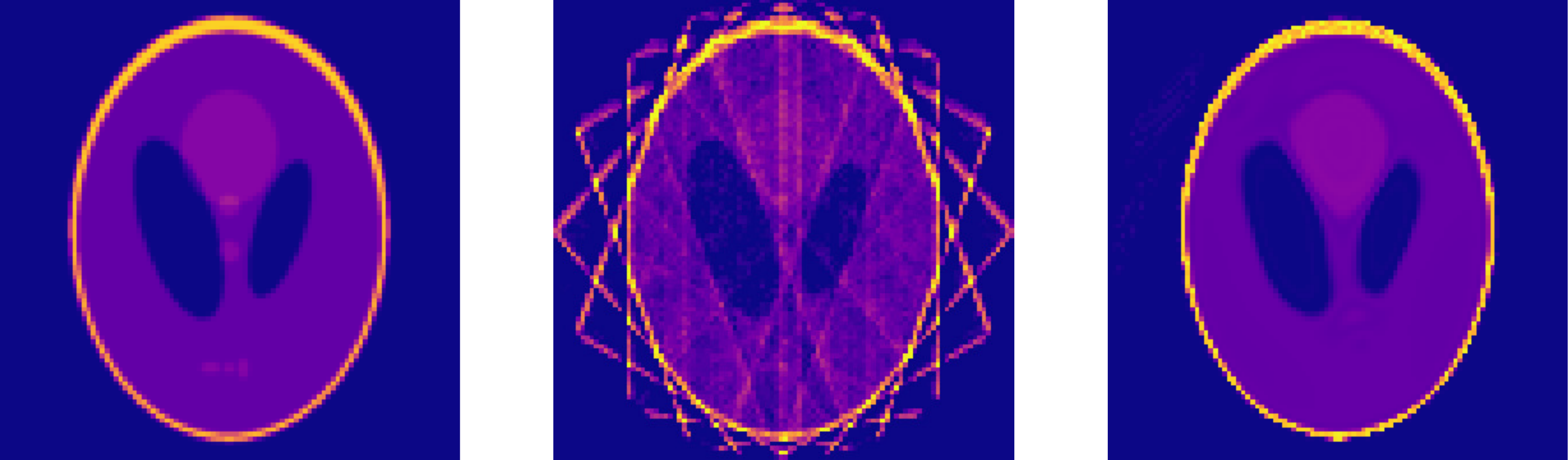}}
		%	\vspace{-2mm}
		\caption{Left: Shepp-Logan phantom. 
			Middle: FBP reconstruction. 
			Right: proposed method.%Reconstruction using our proposed method.
		}
		\label{fig:phantom}
	\end{center}
	\vskip -0.35in
\end{figure}

Figure~\ref{fig:phantom} shows the ground truth image along with the reconstructions obtained with FBP and our proposed method, respectively.
The drawback of the FPB in this case is obvious, as is seen from the distortions present both inside and outside the main ellipse.  
The GP model, on the other hand, is much more homogeneous within the respective regions.
However, some blurriness is observed. 

\subsubsection{Real CT data}
As a real-world example, we consider the carved cheese data set provided by the Finish Inverse Problems Society, freely available and documented online \cite{fips,Bubba}.
The data is down-sampled to contain $15$ projections evenly spaced in $[0,336]^\circ$ with $140$ measurements each, hence $2\,100$ in total.
%In particular, the shape of the carved letters
%makes the target rather challenging for typical sparse and limited angle
%tomography applications \cite{Bubba} (exact cite, rephrase if use)

The result is seen in Figure~\ref{fig:cheese}; the leftmost plot shows a dense FBP reconstruction obtained from the complete set of $360$ projections each with $2\,240$ measurements, that is more than $800\,000$ measurements in total. 
Hence, it is considered close to ground truth.
Regarding the other reconstructions, the performance is similar to what we observed in the previous experiment.   
Clearly, the GP model produces a solution in which the different regions are better distinguished.
There are some parts with blurry elements, primarily near the boundaries of the characters; it is likely that these effects could be overcome by a more well-designed network structure.

\section{Related work}
%With a long and successful history, GP models have been developed for a very broad class of problems.
%Being the most crucial model component, it is no surprise that much work has focused on customisation and modification of the covariance function.
%For instance, the field of constrained GPs has gained a lot of interest throughout the years from many differing perspectives \cite{Abrahamsen2001,DaVeiga2012,Maatouk2017,Salzmann2010ImplicitlyCG,Rudovic_shape-constrainedgaussian,Ginsbourger2017,Frellsen,Thore2003,Nguyen201569,Nguyen201652,Alvarez2012,Constantinescu2013,Jidling2017,Hegermann2018}    

The use of input transformations in the covariance function is by no means a new construction; it is used in modelling solar radiation patterns \cite{Sampson1992nonparametric} and to impose periodicity \cite{MacKay_gp_intro}, which in turn is exploited in modelling of the atmospheric carbon dioxide concentration \cite{Rasmussen2006} and for long-term forecasting \cite{HajiGhassemi14}. 
In \cite{Snelson2006}, a linear input transformation is used for dimensionality reduction in sparse GPs. 
Another closely related approach is to transform the GP \textit{outputs}, which relaxes the embedded Gaussianity assumptions \cite{Snelson2004}.

Incorporation of deep learning into GPs has a long history as well.
The neural network covariance function \cite{Neal1996} is particularly notable, encoding a one-layer neural network with infinitely many neurons.
Another area that has gained a lot of interest in recent years is constituted by the \textit{deep GPs} \cite{Damianou2013DeepGP,DamianouPHD}, where a series of GPs are combined in a network structure.
However, the computational demand is rather intricate; scalable extensions of this model are developed in \cite{Dai2015variational,Salimbeni2017}, with variational inference being a key component.

As we have already mentioned, the foundation that this work relies upon is a technique referred to as manifold GPs \cite{Calandra2016} or deep kernel learning \cite{Wilson16}, both of which describe more or less the same procedure in slightly different contexts.
An interesting extension is found in \cite{Wilson2016stochastic}, where the framework is generalised to a broader class of problems using stochastic variational inference.
Another similar approach considers a model customised for recurrent structures \cite{Shedivat2017}, where the performance using standard covariance functions is poor.
In \cite{Garriga2019}, a low-parameterised relative to deep kernel learning is developed with focus on convolutional neural networks.  
The construction has also gained interest as a potential tool in Bayesian optimisation \cite{Shahriari2016taking,Wu2017}.
Closely related viewpoints are presented in \cite{Lee2018deep,Matthews2018}, which both consider the relation between GPs and wide deep neural networks.

The vast majority of GP models developed are concerned with point measurements;
although integral measurements are not as common, they are present in relevant real-world applications, including CT reconstruction used for demonstration in this work.
The CT problem has been successfully attacked from several different angles, using deep learning techniques \cite{Pelt2018,Hammernik2017,Adler_primal_dual2018,Wurlf_dlct,Adler_solving_ill_posed2017} and statistical methods \cite{Kaipio2006statistical,Sauer1994bayesian,Bouman1996unified,Haario2017shape} including the GP \cite{Purisha_prob_approach}, but not previously with deep kernel learning.
Another area of rising importance is strain field estimation based on the longitudinal ray transform \cite{Lionheart2015,Santisteban2002,Santisteban_Eng}, which constitute a line integral of the projected strain tensor.
Since it involves the reconstruction of a multidimensional function, it is a technically more challenging problem than the CT equivalent.
GPs tailored to satisfy the physical constraints of the strain field have been used to this end \cite{Jidling2018,Hendriks2017}, but so far no deep learning based techniques; the proposed method serves as an interesting extension.  
Yet another example of integral measurements in GPs are found within stochastic optimisation \cite{Hennig2013}, where the secant condition used in quasi-Newton methods is replaced by its exact counterpart; this approach has shown promising results in nonlinear system identification \cite{Wills2017}.   

\section{Conclusion and future work}
In this work we have presented a method that applies deep kernel learning to problems with integral measurements.
We proposed utilising a basis function expansion to make the computations practically feasible, and pre-training of the neural network to improve the result of the joint parameter training.
The method was illustrated on both simulated and real data from X-ray computed tomography, indicating a promising potential.

Future work may focus on customisation of the neural network structures. 
Having paid a fairly limited attention to this important part of the model, we believe that the room for improvement is significant.
Moreover, although the neural network is one possible choice of latent mapping, it is by no means the only one.
Other alternatives are also worth exploring, as well as their potential combinations. 
As mentioned in Section \ref{sec:training} we did restrict ourselves to a single latent output to reduce the computational burden; however, a well-designed combination of several outputs with differing mappings is likely to improve the performance.
Also, extensions of deep kernel learning aimed at reducing the risk of over-fitting should be explored to further improve the robustness. 

\section{Acknowledgements}
This research was financially supported by the Swedish Foundation for Strategic Research (SSF) via     the project \emph{ASSEMBLE} (contract number: RIT15-0012).

\section{Supplementary material}
\subsection{Selecting the domain size}\label{app:select_Lk}
Here we discuss the selection of the domain size of $\Omega$, which is determined by the parameters $L_k$ used in building the basis functions \eqref{eq:basis_funcs}.
A basic requirement is that $L_k$ should be clearly larger than the maximum absolute size of $u_k(\x)$ to avoid undesired impact of the Dirichlet conditions used in the eigenvalue problem \eqref{eq:Laplace}.  
Apart from this, the size of $L_k$ determines the quality of the approximation specified in the frequency domain; given a fixed number of basis function in direction $k$, increasing $L_k$ yields a higher frequency resolution $\frac{\pi}{2L_k}$ in that direction, while at the same time it is reducing the frequency range $[\frac{\pi}{2L_k},\,\,\frac{\tilde{m}\pi}{2L_k}]$. 
A reasonable approach is to select $L_k$ with respect to the spectral frequency $S(\cdot)$, so that the domain covers the vast majority of the spectral "mass" (equivalently to how confidence regions covers different amounts of the probability mass). 
This is dependent on the lengthscale parameters, and we suggest selecting $L_k$ such that $\alpha l_k^{-1} = \max_k c_{kj}$,
where the parameter $\alpha$ is chosen with respect to the spectral density of the covariance function used.
For instance, the value $\alpha=5$ is reasonable for the squared exponential covariance function \eqref{eq:se} and yields a coverage of more than $99.9\%$.
Note that since $L_k$ is not part of the optimised parameters, the recalculation modifies the definition of the cost function.
However, that effect is negligible and the strategy has proven very useful in practise.

\subsection{Numerical Implementation} % (fold)
\label{app:numerical_robustness}
The numerical robustness can be improved using the $QR$-factorisation, considering both the computations of the loss function and the predictions. 
For instance, the LOO-CV procedure requires $(\Phi\Lambda\Phi^\Transp+\sigma^2I)^{-1}$ and $(\Phi\Lambda\Phi^\Transp+\sigma^2I)^{-1}\y$.
To that end, we first compute the matrix $R$ in the $QR$-factorisation
\begin{align}
QR=
\begin{bmatrix}
\Lambda^{1/2}\Phi^\Transp\\
\sigma I
\end{bmatrix}.
\end{align} 
Since $Q$ is a unitary matrix, it follows that $R^\Transp R= \Phi^\Transp\Lambda\Phi+\sigma^2I$ and so the desired quantities can be found using efficient forward and backward substitutions \cite{golubloan}.

Since PyTorch's automatic differentiation is being used to provide the partial derivatives of the cost function with respect to the parameters $\theta$, a `backwards' method is required for the QR-factorisation. 
Although a QR algorithm is implemented in PyTorch, it does not have a backwards method in its current stable release. 
Given the partial derivative of the cost function $C$ with respect to $R$, the backwards algorithm needs to compute the partial derivates of the cost function with respect to the elements of $A$, where $QR = A$.
Algorithm~\ref{alg:backwards_qr} provides a backwards method that can be added to the QR function in PyTorch and is based upon the equations presented in \cite{WalterPHD}.

\begin{algorithm}[ht]
	\caption{Backwards Method for QR}
	\label{alg:backwards_qr}
	\begin{algorithmic}
		\STATE {\bfseries Input:} $\frac{\partial C}{\partial R}$, $Q$, and $R$
		\STATE {\bfseries Output:} $\frac{\partial C}{\partial A}$
		\STATE {\bfseries 1:} Compute the psuedoinverse of $R$: $R^+ = (R^\top R)^{-1}R^\top$ 
		\STATE {\bfseries 2:} Compute $\beta = \left(R\frac{\partial C}{\partial R} - \frac{\partial C}{\partial R} R^\top\right)$
		\STATE {\bfseries 2:} Extract the lower triangular matrix below the main diagonal: $\Gamma = \text{tril}(\beta,-1)$
		\STATE {\bfseries 3:} Compute the output: $\frac{\partial C}{\partial A} = Q(\frac{\partial C}{\partial R} + \Gamma R^{+\top})$ 
	\end{algorithmic}
\end{algorithm}

% references
%\red{References: full first names, middle names shortened, polish info}
\bibliography{arxiv.bib}
\bibliographystyle{plain}
	
\end{document}